# Classification of Deceased Patients from Non-Deceased Patients using Random Forest and Support Vector Machine Classifiers

CS-521 Project Report

Biraj Tiwari (tbiraj@unm.edu)
Dheeman Saha (dsaha@unm.edu)
Aaron Segura (AaSegura@salud.unm.edu)

**Abstract**

Analyzing large datasets and summarizing it into useful information is the heart of the data mining process. In healthcare, information can be converted into knowledge about patient historical patterns and possible future trends. During the COVID-19 pandemic, data mining COVID-19 patient information poses an opportunity to discover patterns that may signal that the patient is at high risk for death. COVID-19 patients die from sepsis, a complex disease process involving multiple organ systems. We extracted the variables that physicians are most concerned about regarding viral septic infections. With the aim of distinguishing COVID-19 patients who survive their hospital stay and the COVID-19 that do not, the authors of this study utilize the Support Vector Machine (SVM) and the Random Forest (RF) classification techniques to classify patients according to their demographics, laboratory test results, and preexisting health conditions. After conducting a 10-Fold Cross Validation procedure, we assessed the performance of the classification through a Receiver Operating Characteristic (ROC) curve and a Confusion Matrix was be used to determine the accuracy of the classifiers. We also performed a cluster analysis on the binary factors such as if the patient had a preexisting condition and if sepsis was identified, as well as the numeric values from patient demographics and laboratory test results as predictors.

## Introduction

COVID-19 causes sepsis, an overwhelming and life-threatening response to an infection leading to organ damage and death. Patients can develop sepsis from many different types of infections, the most common being pneumonia. In pneumonia, the patient suffers from severe lung inflammation in response to an infection where the air sacs fill with pus causing shortness of breath and inducing heart strain due to the lack of oxygen. Bacteria, viruses, and fungus can all cause the infection causing pneumonia and therefore sepsis increasing the patient's risk of suffering permanent organ damage or death.

In the case of COVID-19, the virus affects the multiple organ systems in the body, and its relationship to sepsis is still not completely understood. Thus, making it more difficult for researchers to determine the effectiveness of care in sepsis patient populations and understand the impact the SARS-CoV-2 virus has had on the sepsis patient population and contributing to patient death. Therefore, it is essential for hospitals and healthcare workers to properly identify COVID-19 patients with sepsis and assess these in a timely manner as one in three patients who die in hospitals die from sepsis [1].

Now with COVID-19 patients contributing to the overall sepsis patient population, it is important to determine which patient attributes contribute to the likelihood of the patient surviving their stay in the hospital. We considering the patient's initial and last lab values in their hospital stay as well as the patient's preexisting conditions prior to coming into the hospital to develop a model for the patient likelihood of survival. These factors are crucial in the treatment response in caring for sepsis patients as patients improve or decline during their stay in the hospital. We also include in our analysis the patient demographics such as age and gender, race, ethnicity, body height, and body weight, as well as noting if sepsis was identified.

Using the national database of COVID-19 patients, we will model the important factors such as laboratory values and patient comorbidities to obtain the patient's effective classification of patients due to COVID-19.

Furthermore, our results could be used for predicting the COVID-19 patient population survival rate as well as the potential of developing sepsis which could then be used for comparing to the non-COVID-19 sepsis population. Lastly, we may aid in determining areas of opportunity that can help to reduce the overall COVID-19 sepsis mortality rate and may potentially direct future research activities.

## Dataset Explanation and Analysis

Our efforts are in coordination with UNMH clinical researchers Drs. Jon Femling, MD, PhD and Rahul Shekhar, both have provided external guidance and support on our research activities namely identifying which parameters to consider in our analysis and which patient conditions might be relevant to consider in regards to the COVID-19 patient population. Dr. Shekhar is a hospitalist and clinical education professor in the department of internal medicine and Dr. Femling is a physician scientist in the departments of microbiology and emergency medicine.

After obtaining IRB approval, listing the aforementioned as Principal /Co-Investigators, we obtained access to Cerner AWS COVID-19 data cohort database on 10/25/2020 (IRB UNM-HSC project ID 20-607 COVID-19 and Sepsis). We were allowed access to HealtheDataLab$^{TM}$ – the Cerner Data Science Ecosystem, built and deployed on Amazon Web Services (AWS). Cerner is only offering health systems free access to a COVID-19 data science workspace for academic-funded projects thus it was required that we submit IRB and data request proposals to the UNM-HSC IRB and Cerner, respectively.

A Jupyter notebook interface with Spark infrastructure was provided by the Cerner Learning Health Network initiative and the data user agreement prevents us from releasing any real data however we can present the results of our analyses. Using the Spark infrastructure, we primarily imported PySpark and Python 3 libraries to preprocess the data from the Cerner database. We have 7 prominent tables, however, for the purposes of this paper we only consider the following 4 tables:

> Table 1) **Demographics** - information about the patient, their demographics information, and their deceased/alive status
> **personid**: The ID of the person
> **gender**: The gender of the person
> **race**: The race of the person
> **ethnicity**: The ethnicity of the person
> **deceased** status: An indicator of the death of the person
>
> Table 2) Covid_labs - Includes qualitative COVID lab results associated with both qualifying and supplemental encounters. Additionally, the table includes any results from up to two weeks prior to the service date of a qualifying encounter, where no positive result was identified on the qualifying encounter itself.
> **personid**: The ID of the person associated with the result
> **result**: The value of the lab. Possible values include Positive, Negative, Indeterminate, Not Done, or Unknown
> **pos_cvd19_lab_ind**: A flag indicating a positive result that is generally consistent with COVID-19 infection.
>
> Table 3) **Result** - For each qualifying patient, includes all result records from encounters with service dates on or after 1/1/2015.
> **personid**: The ID of the person associated with the result
> **result**: The display name of the test or measurement
> **numericvalue**: The nsumeric value of the result.
>
> Table 4) **Condition** - For each qualifying patient, includes all available diagnosis records from encounters with service dates on or after 1/1/2015.
> **personid**: The ID of the person associated with the condition
> **codetype**: The type of coding system used for recording the condition
> **conditioncode**: The code value that identifies the condition, for example, an ICD-10-CM code
> **condition**: The display name of the condition

We began our analysis with the following steps:
1. First, we extracted the personids from the database such that we obtained 50% deceased and non-deceased patients.
2. Then, for each of personid we chose parameters of interest per Dr. Shekhar's guidance from the result and condition tables to obtain the vitals, body measurements, laboratory results and patient preexisting conditions, respectively. We extracted this information through iterations of 1000 patients at a time. We limited the extraction iteration count to 1000 due to memory constraints posed by Cerner storing stored the extracted data in CSV files. After obtaining each CSV files for all the result and condition tables with then merged the datasets into a master CSV file containing all the patient information. We obtained a list of patients who were dead and balanced the data that consisted of equal number of dead/alive patients thus causing us to drop non-deceased patients from our analysis. We finally, obtained data for 9366 patients with 227 columns consisting of 4683 deceased and 4683 non-deceased patients.
3. Since we have data for each personid with service dates on or after 1/1/2015, we restricted to the COVID-19 positive patients as indicated by their laboratory status and the year 2020.

4. We selected the first and last laboratory values and then pivoted the row into columns by results for each personid and created two different two different tables each for the first value and last value. We consider only the first lab values for our analysis here to classify the patients who have died with COVID-19. Below is an example of the pivoting process:

Before:

| personid | result | Numeric value |
|---|---|---|
| A | Body weight | 67 |
| A | Age − Reported | 51 |
| A | Leukocytes [#/volume] First.. | 23 |
| A | Leukocytes [#/volume] Last.. | 46 |
| A | Respiratory rate | 43 |

After:

| personid | Body weight | Age Reported | Leukocytes [#/volume] First | Leukocytes [#/volume] Last | Respiratory rate |
|---|---|---|---|---|---|
| A | 67 | 51 | 23 | 46 | 43 |

Table 1: Tables Before and After Pivoting our data

5. For our laboratory result data, we chose the attributes with more than 50% response rate and addressed the missing values by sampling randomly from a normal distribution of the z-score non-missing values to replace the missing values for each parameter (see Algorithm 1).

   1. First, we excluded negative values since it is nonsensical to have measured negative measurements.
   2. Then, we standardized the data by subtracting the minimum from each value of the data from that point and then dividing by the difference between the maximum and minimum for the numeric data.
   $$standardized\ value = \frac{value - \min(attribute)}{\max(attribute) - \min(attribute)}$$
   3. The data we obtained after preprocessing consists of values ranging between 0 and 1 for the numeric data and the binary data remained unchanged 0 and 1 after the standardization. In other words, for the condition table, we filled the not applicable values with zero for patients who did not have the condition such that we have only binary values for each entry regarding the patient's comorbidities.

```
Algorithm 1 Resample: ColumnVector
 1: for iteration = 1, 2, . . . NumCols do
 2:     Drop the NAs from each column
 3:     Calculate the Z − Score
 4:     for iteration = 1, 2, . . . NumCols do
 5:         Normal Dist. value on Z − Score Mean and Standard Deviation
 6:         while value < 0 do
 7:             Re-Calculate Normal Dist. value again
 8:         end while
 9:         Check for NAs
10:         if Index Value == NA then
11:             ColumnVector[iteration] = value
12:         end if
13:     end for
14:     return ColumnVector
15: end for
```

Algorithm 1: Resampling from Normal Distribution for Missing Values (NaN)

6. We then reduced dimensionality with a Principal Component Analysis on the numeric values and used the first two principal components for data visualization.
7. A Hierarchical clustering of both the binary and numeric data was performed followed by a Hierarchical clustering of the numeric data only
8. We used Support Vector Machine (SVM) for classification of the numeric data separating the deceased and non-deceased patients with 10-fold cross validation
9. Random Forest classification of the numeric data separating the deceased and non-deceased patients with 10-fold cross validation obtaining accuracy
10. DBSCAN of the numeric data identifying two clusters
11. Confusion matrices for both Random Forest and SVM.
12. ROC (Receiver Operating Characteristic) curves plotting the True Positive Rate vs False Positive Rate for the results of the Random Forest classifier.
13. A statistical outlier detection was performed identifying points either $< -3\sigma$ or $> 3\sigma$ as well as boxplots that identify outliers for data point that are $< median - 1.5 \times IQR$ and $> median + 1.5 \times IQR$. A Local Outlier Factor (LOF) which also serves to identify outliers in the dataset was plotted for each of our points.

Below are plots of a few rows of the data and preliminary statistics are provided for our most important features in the Appendix (Tables 99 & 99).

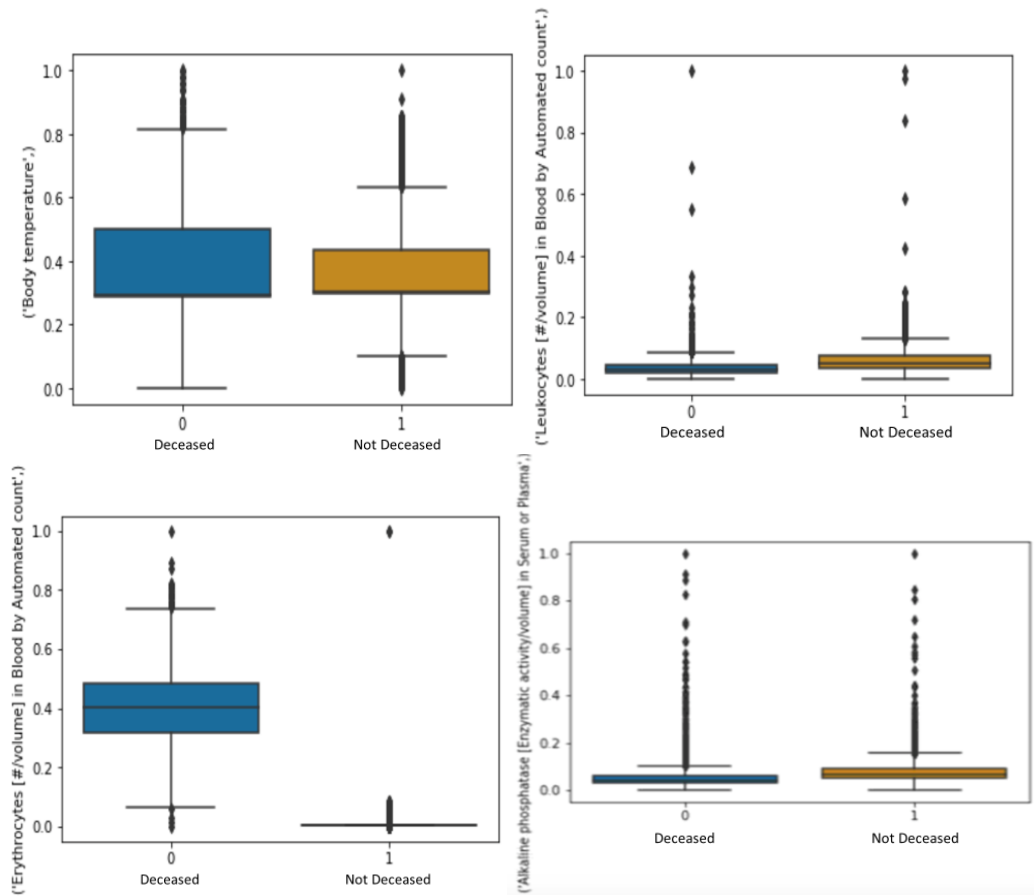

Figure 1: Four Boxplots of Numeric Attributes showing the spread of the standardized data between 0 and 1 for Body Temperature (Upper Left), Leukocytes [#/Volume] in Blood by Automated count (Upper Right), Erythrocytes [#/volume] in Blood by Automated count (Lower Left), and Alkaline phosphatase [Enzymatic activity/volume] in Serum or Plasma (Lower Right) by Deseased and Not Deceased patients.

In Figure 1, we have four boxplots of numeric attributes of our data where 0 indicates deceased patients and 1 indicates that the patient was alive or not deceased. Notice that this is after we have normalized the data, scaling from 0 to 1. In the boxplots, we can see variations in the spread in our data. In the upper left boxplot, we see roughly normally distributed data with outliers in the Not deceased boxplot since these points are more than $1.5 \times IQR$ from the median. In the upper and lower right boxplots, we see that the data is right skewed for both the deceased and not deceased patients as well as a high number of outliers. In the lower left boxplot, we see that the deceased patients are more normally distributed whereas the patients that not deceased have a right skewed distribution for this attribute.

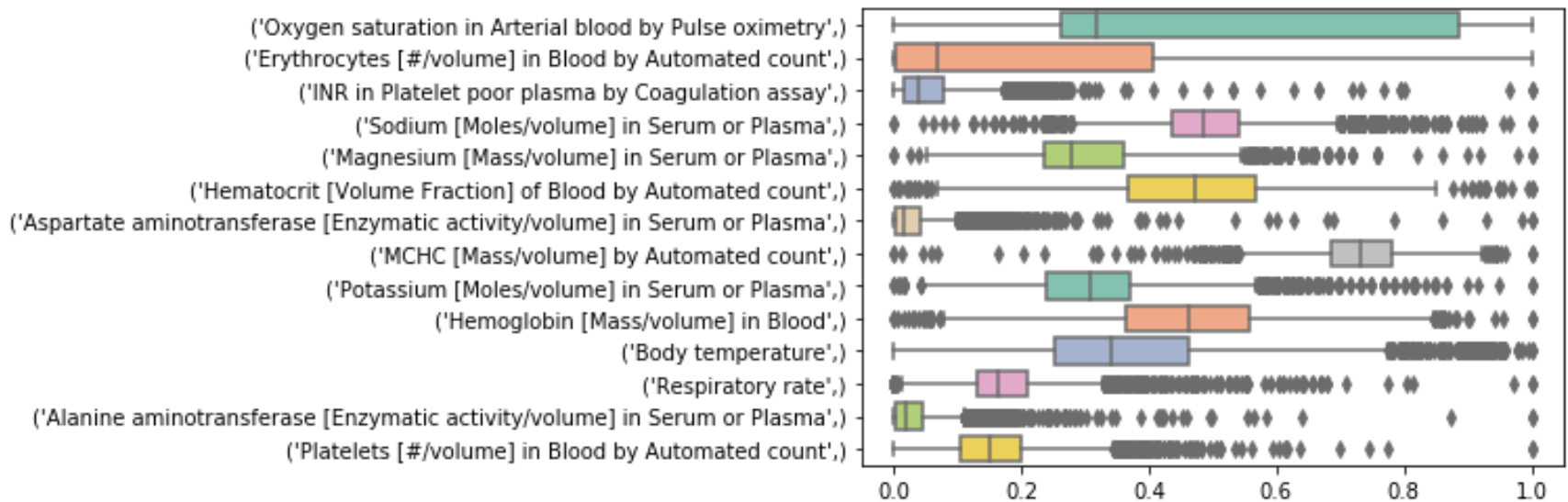

Figure 2: Boxplots of 14 Most Important Numeric Features in classifying deceased and non-deceased COVID-19 patients of the standardized data between 0 and 1 ranked from most important to least most important, Oxygen saturation in Arterial blood by Pulse oximetry to Platelets [#/volume] in Blood by Automated count, respectively,

In Figure 2 above, we see a more comprehensive group of boxplots of our data which we will discuss later that these are the most important numeric features that correctly classify deceased and non-deceased COVID-19 patients. Notice that we have a significant number of outliers in these boxplots which we will address below using statistical analyses as well as LOF.

**Algorithms and Hyperparameters**

Initially, there were over 200 attributes that are included in the dataset we extracted, however, considering all the attributes for the classification purpose might be too computationally expensive. Thus, we have decided to perform feature selection for classification purposes. For the feature selection purpose, we used the ExtraTreeClassifier, which resembles the same learning method as that of a Decision Tree. This type of tree classifier randomizes the splitting decision and subset the data to reduce over-learning from the given dataset and overfitting.

The splitting criteria are based on the value of the Gini Index, which varies between 0 and 1. The result 0 represents the purity of the classification whereas the value 1 represents randomized values among the elements. Finally, using the ExtraTreeClassifier, we ranked all the features in the descending order based on their feature importance value. From the outcome we have selected the top 20 features as the values of the discarded features are extremely small. Below we have included information pertaining to the gini index, $gini(D)$, which is defined as

$$gini(D) = 1 - \sum_{j=1}^{n} p_j^2$$

Where $p_j$ is the relative frequency of class $j$ in $D$. Now if a dataset $D$ is split on $A$ into two subsets $D_1$ and $D_2$, then the gini index $gini(D)$ will be defined as

$$gini_A(D) = \frac{|D_1|}{|D|} gini(D_1) + \frac{|D_2|}{|D|} gini(D_2)$$

where the reduction in impurity being written as $\Delta gini(A) = gini(D) - gini_A(D)$. This attribute provides the smallest $gini_{split}(D)$ mentioned before. We present the results of this algorithm in Table 3 and Figure 3.

Regarding outlier detection, we computed the Local outlier factor (LOF) which uses reachability distance. Reachability distance from $o'$ to $o$ is defined as $reachdist_k(o \leftarrow o') = \max\{dist_k(o), dist(o, o')\}$ where $k$ is a user-specified parameter. Now the local reachability density of $o$ or $lrf(o)$ is:

$$lrd(o) = \frac{||N_k(o)||}{\sum_{o' \in N_k(o)} reachdist_k(o' \leftarrow o)}$$

And the LOF of an object $o$ is the average of the ratio of local reachability of $o$ and those of $o$'s $k$-nearest neighbors as shown in the following formula:

$$LOF_k(o) = \frac{\sum_{o' \in N_k(o)} \frac{lrd_k(o')}{lrd_k(o')}}{||N_k(o)||} = \sum_{o' \in N_k(o)} lrd_k(o') \cdot \sum_{o' \in N_k(o)} reachdist_k(o \leftarrow o)$$

Note that the lower the local reachability density of $o$, and the higher the local reachability density of the KNN of $o$, the higher LOF.

Another measure we used to evaluate our clustering is the Silhouette Coefficient which is calculated by using the mean intra-cluster distance $a$ and the mean nearest-cluster distance $b$ for each sample. The Silhouette Coefficient for a sample is $(b - a) / max(a, b)$. To clarify, b is the distance between a sample and the nearest cluster that the sample is not a part of making the Silhouette coefficient an internal measure of the clustering. Below are the formulas for calculating $a$, $b$, and $s$ for object $o$:

$$a(o) = \frac{\sum_{o' \in C_i, o \neq o'} dist(o, o')}{||c_i|| - 1}, b(o) = \min_{C_j : 1 \leq j \leq k, i \neq j} \frac{\sum_{o' \in C_i} dist(o, o')}{||c_i||}, s(o) = \frac{b(o) - a(o)}{\max(a(o), b(o))}$$

Note that the Silhouette metric is a distance calculation algorithm using Euclidean or Manhattan distance and the Silhouette Score ranges between -1 to 1. Here, a high Silhouette score suggests that the objects are well matched to their own cluster and poorly matched to their neighborhood clusters.

| Value | Interpretation |
|---|---|
| 0.71-1.0 | A strong structure has been found |
| 0.51-0.70 | A reasonable structure has been found |
| 0.26-0.50 | The structure is weak and could be artificial. Try additional methods of data analysis. |
| <0.25 | No substantial structure has been found |

Table 2: Silhouette Coefficient Metric Value with Interpretation

Below we have included the Pseudocodes for the algorithms used to obtain our results.

Pseudocode for Random Forest Algorithm

**Precondition:** A set $S$ of objects $S := (x_1, y_1), \ldots, (x_n, y_n)$, features $F$, and number of trees in forest $B$

```
     function RandomForest(S, F):
1      H ← ∅
2      for i ∈ 1, …, B do
3        S^(i) ← A bootstrap sample from S
4        h_i ← RandomizedTreeLearn(S^(i), F)
5        H ← H ∪ {h_i}
6      end for
7      return H
9    end function
10   function RandomizedTreeLearn(S, F):
11     At each node:
12       f ← very small subset of F
13       Split on best feature in f
14     return The learned tree
15   end function
```

Algorithm 2: Random Forest Algorithm

Pseudocode for Hierarchical clustering Algorithm

**Precondition:** A set $X$ of objects $\{x_1, \ldots, x_n\}$
A distance function $dist(c_1, c_2)$
**function** $HeirarchicalCluster(D, eps, MinPts)$:
1    **for** $i = 1$ to $n$
2       $C = \{x_i\}$
3       end for
4    $C = \{c_1, \ldots, c_n\}$
5    $I = n + 1$
6    **while** $C.size > 1$ **do**
7       $(c_{min1}, c_{min2}) = min_{dist}(c_i, c_j)$ for all $c_i, c_j$ in $C$
8       remove $c_{min1}$ and $c_{min2}$ from $C$
9       add $\{c_{min1}, c_{min2}\}$ to $C$
10     $I = I + 1$
11  **end while**
12 **end function**

Algorithm 3: Hierarchical clustering Algorithm

Pseudocode for DBSCAN Algorithm

**Precondition:** A dataset $D$ containing
  **function** $DBSCAN(D, eps, MinPts)$:
1    $C = 0$
2    **for** each unvisited point $P$ in dataset $D$
3       mark $P$ as visited
4       $NeighborPts = regionQuery(P, eps)$
5       **if** $sizeof(NeighborPts) < MinPts$
6       mark $P$ as $NOISE$
7       **else**:
8          $C = next\ cluster$
9          $expandCluster(P, NeighborPts, C, eps, MinPts)$
10      add $P$ to cluster $C$
11     **for** each point $P'$ in $NeighborPts$
12        **if** $P'$ is not visited
13          mark $P'$ as visited
14          $NeighborPts' = regionQuery(P', eps)$
15        **if** $sizeof(NeighborPts') >= MinPts$
16          $NeighborPts = NeighborPts$ joined with $NeighborPts'$
17        **if** $P'$ is not yet member of any cluster
18          Add $P'$ to cluster $C$
19      $regionQuery(P, eps)$
20        **return** all points within $P$'s eps- neighborhood (including $P$)
21 **end function**

Algorithm 4: DBSCAN Algorithm

Pseudocode for Support Vector Machine (SVM) Algorithm

**Output:** Optimal Value for $C$ and $\gamma$
**Input:** $p, InitArhiveSize, Growth, MaxArchiveSize, MaxStagIter, m$, and termination criterion

```
1    function SVM(p, InitArhiveSize, Growth, MaxArchiveSize, MaxStagIter, m):
2        k = InitArhiveSize
3        initialize k solutions
4        call SVM algorithm to evaluate k solutions
5        while classification accuracy 100% or number of iteration ≠ 10 do
6          if rand(0,1) < p then do
7            for i = 1 to no. of iterations do
8              select best selected solution
9              Sample best selected solution
10                Call SVM algorithm to evaluate the new generated solutions
11                   if Newly generated solution is better than Sbest
12                        Substitute for Sbest
13                   end
14            end
15               else
16                   for j = 1 to k do
17                       Select Sbest selected
18                   end
19               end
20             end
21        else:
22           for j = 1 to k do
23              Select S
24              Sample selected S
25              Store newly generated solutions
26              Call SVM algorithm to evaluate the new generated solutions
27              if Newly generated solution is better than S_j then
28                   Substitute newly generated solution for S_j
29              end
30           end
31           if current iterations are multiple of Growth & k < MaxArchive Size then
32              Initialize new solution
33              Add new solution to the archive
34              k++
35           end
36           if # (number) of iterations without improving classification accuracy of Sbest = MaxStagIter
37              if # (number) of iterations without improving classification accuracy of Sbest = MaxStagIter
38                   Re- initialize T (solution archive) but keeping Sbest
39              end
40           end
41    end function
```

Algorithm 5: Support Vector Machine (SVM) Algorithm

## Tables of Hyperparameters used by Algorithm

| **Random Forest Classifier** Hyperparameter | Value | Description/Explanation |
|---|---|---|
| Bootstrap | False | Whether bootstrap samples are used when building trees. If False, the whole dataset is used to build each tree. |
| max_depth | 50 | The maximum depth of the tree. If None, then nodes are expanded until all leaves are pure or until all leaves contain less than min_samples_split samples |
| n_estimatorsint: | 2000 | The number of trees in the forest. |
| min_samples_leaf | 2 | The minimum number of samples required to be at a leaf node. A split point at any depth will only be considered if it leaves at least min_samples_leaf training samples in each of the left and right branches. |
| min_samples_split: | 10 | The minimum number of samples required to split an internal node: |

| **Random Forest Regressor** Hyperparameter | Value | Description/Explanation |
|---|---|---|
| bootstrap | False | Whether bootstrap samples are used when building trees. If False, the whole dataset is used to build each tree. |
| ccp_alpha | 0.0 | Complexity parameter used for Minimal Cost-Complexity Pruning. The subtree with the largest cost complexity that is smaller than ccp_alpha will be chosen |
| criterion | 'mse' | The function to measure the quality of a split. Supported criteria are "mse" for the mean squared error, which is equal to variance reduction as feature selection criterion, and "mae" for the mean absolute error. |
| max_depth | 30, | The maximum depth of the tree. If None, then nodes are expanded until all leaves are pure or until all leaves contain less than min_samples_split samples. |
| max_features | 'auto', | The number of features to consider when looking for the best split: |
| max_leaf_nodes | None | Grow trees with max_leaf_nodes in best-first fashion. Best nodes are defined as relative reduction in impurity. If None then unlimited number of leaf nodes. |
| max_samples | None | If bootstrap is True, the number of samples to draw from X to train each base estimator. |
| min_impurity_decrease | 0.0 | A node will be split if this split induces a decrease of the impurity greater than or equal to this value. |
| min_impurity_split | None, | Threshold for early stopping in tree growth. A node will split if its impurity is above the threshold, otherwise it is a leaf |
| min_samples_leaf | 4 | The minimum number of samples required to be at a leaf node. A split point at any depth will only be considered if it leaves at least min_samples_leaf training samples in each of the left and right branches. This may have the effect of smoothing the model, especially in regression. |
| min_samples_split | 2 | The minimum number of samples required to split an internal node: |
| min_weight_fraction_leaf | 0.0 | The minimum weighted fraction of the sum total of weights (of all the input samples) required to be at a leaf node. Samples have equal weight when sample_weight is not provided. |
| n_estimators | 2000 | The number of trees in the forest |
| verbose | 0 | Controls the verbosity when fitting and predicting. |
| warm_start | False | When set to True, reuse the solution of the previous call to fit and add more estimators to the ensemble, otherwise, just fit a whole new forest. |
| n_jobs | None, | The number of jobs to run in parallel |
| oob_score | False, | whether to use out-of-bag samples to estimate the R^2 on unseen data |

| random_state | 42 | Controls both the randomness of the bootstrapping of the samples used when building trees |

**Density-Based Spatial Clustering of Applications with Noise (DBSCAN)**

| Hyperparameter | Value | Description/Explanation |
|---|---|---|
| Eps | i | The maximum distance between two samples for one to be considered as in the neighborhood of the other. This is not a maximum bound on the distances of points within a cluster. This is the most important DBSCAN parameter to choose appropriately for your data set and distance function. |
| min_samples | 10 | The number of samples (or total weight) in a neighborhood for a point to be considered as a core point. This includes the point itself. |

**Support Vector Machine (SVM)**

| Hyperparameter | Value | Description/Explanation |
|---|---|---|
| C | 0.01 | If you have a lot of noisy observations, you should decrease it: decreasing C corresponds to more regularization. |
| break_ties | false | T If true, decision_function_shape='ovr', and number of classes > 2, predict will break ties according to the confidence values of decision_function; otherwise the first class among the tied classes is returned. Please note that breaking ties comes at a relatively high computational cost compared to a simple predict. |
| cache_size | 200, | Specify the size of the kernel cache (in MB) |
| class_weight | None | Set the parameter C of class i to class_weight[i]*C for SVC. |
| coef0 | 0.0 | Independent term in kernel function. It is only significant in 'poly' and 'sigmoid'. |
| decision_function_shape | 'ovr', | Whether to return a one-vs-rest ('ovr') decision function of shape (n_samples, n_classes) as all other classifiers, or the original one-vs-one ('ovo') decision function of libsvm which has shape (n_samples, n_classes * (n_classes - 1) / 2). |
| Degree | 3 | Degree of the polynomial kernel function ('poly'). Ignored by all other kernels |
| Gamma | 1 | Kernel coefficient for 'rbf', 'poly' and 'sigmoid'. |
| Kernel | 'rbf' | Kernel coefficient for 'rbf', 'poly' and 'sigmoid' |
| max_iter | -1 | Hard limit on iterations within solver, or -1 for no limit |
| Probability | False | Whether to enable probability estimates. This must be enabled prior to calling fit |
| random_state | None | Controls the pseudo random number generation for shuffling the data for probability estimates. Ignored when probability is False. Pass an int for reproducible output across multiple function calls |

**Hierarchical Clustering**

| Hyperparameter | Value | Description/Explanation |
|---|---|---|
| color_threshold | 3 | Colors all the descendent links below a cluster node the same color if is the first node below the cut threshold. All links connecting nodes with distances greater than or equal to the threshold are colored with de default matplotlib color 'C0'. If is less than or equal to zero, all nodes are colored 'C0'. If color_threshold is None or 'default', corresponding with MATLAB(TM) behavior, the threshold is set to $0.7 \times max(Z[:,2])$. |
| count_sort | True | 'descending'; the child with the maximum number of original objects in its cluster is plotted first. |
| distance_sort bool | True | For each node n, the order (visually, from left-to-right) n's two descendent links are plotted is determined by this parameter, which can be any of the following values: 'descending' The child with the maximum distance between its direct descendents is plotted first. |

Table 3: Tables of Hyperparameters used by Algorithm

**Experimental Results**

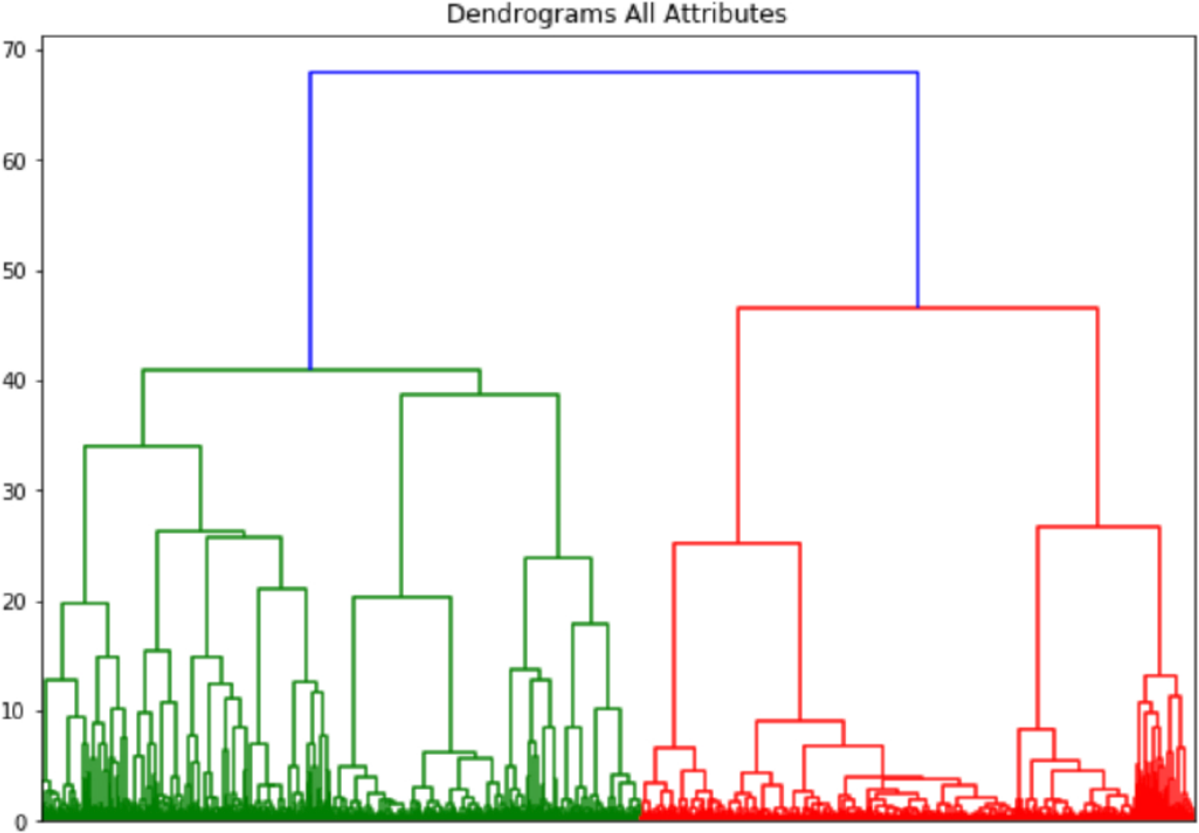

Figure 3: Dendrogram of Hierarchical Clustering for All 59 Attributes
with Deceased Patients (Red) and Non-Deceased Patients (Green)

In Figure 3, we see a Dendrogram of Hierarchical Clustering for All 59 Attributes with Deceased Patients (Red, right) and Non-Deceased Patients (Green, left). We see that we have two major clusters and at about the level 46 we see that the deceased cluster breaks off into two subclusters followed by the non-deceased cluster breaking off into two subclusters around level 41.

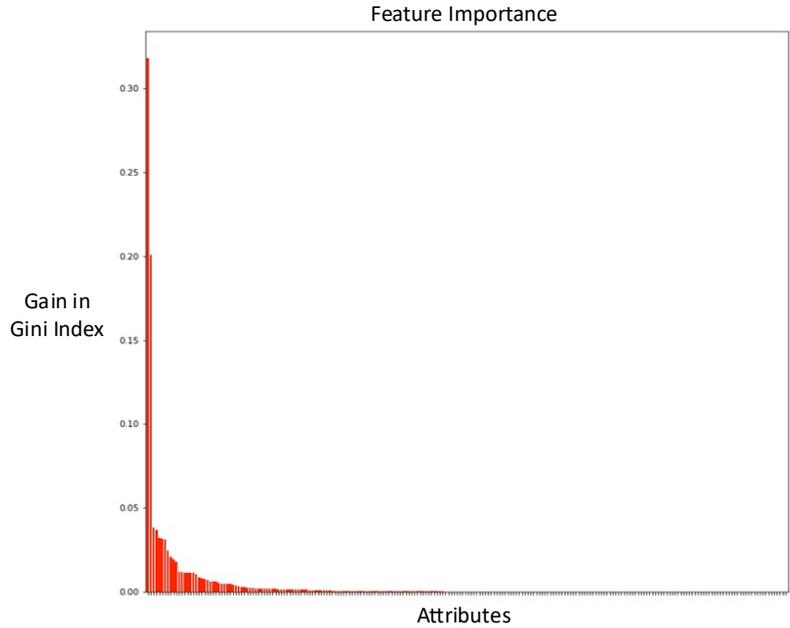

Figure 3: Most Important Features by Gain in Gini Index for Each 227 Attributes

| Rank | Most Important Features Predicting Death from COVID-19 | Data Type |
|------|---------------------------------------------------------|-----------|
| 1 | Oxygen saturation in Arterial blood by Pulse oximetry | Numeric |
| 2 | Erythrocytes [#/volume] in Blood by Automated count | Numeric |
| 3 | Acute kidney failure, unspecified | Binary |
| 4 | INR in Platelet poor plasma by Coagulation assay | Numeric |
| 5 | Severe sepsis with septic shock | Binary |
| 6 | (Sodium [Moles/volume] in Serum or Plasma | Numeric |
| 7 | Magnesium [Mass/volume] in Serum or Plasma | Numeric |
| 8 | Cardiac arrest cause, unspecified | Binary |
| 9 | Hematocrit [Volume Fraction] of Blood by Automated count | Numeric |
| 10 | Aspartate aminotransferase [Enzymatic activity/volume] in Serum or Plasma | Numeric |
| 11 | MCHC [Mass/volume] by Automated count | Numeric |
| 12 | Potassium [Moles/volume] in Serum or Plasma | Numeric |
| 13 | Acute kidney failure with tubular necrosis | Binary |
| 14 | Hemoglobin [Mass/volume] in Blood | Numeric |
| 15 | Body temperature | Numeric |
| 16 | Respiratory rate | Numeric |
| 17 | Acute respiratory distress syndrome | Binary |
| 18 | Alanine aminotransferase [Enzymatic activity/volume] in Serum or Plasma | Numeric |
| 19 | Platelets [#/volume] in Blood by Automated count | Numeric |
| 20 | Metabolic encephalopathy | Binary |

Table 4: Rank of the Top 20 Most Important Features Classifying Deceased COVID-19 Patients by Data Type

From Table 4 and Figure 3, out of the top 20 most important features of 227, we see that the five most important features which classify whether a patient will die from a COVID-19 infection in the dataset are Oxygen saturation in Arterial blood by pulse oximetry, Erythrocyte count in Blood, Acute kidney failure, INR, and Severe sepsis with septic shock. These results were obtained by computing the gain in Gini index to select the most important features in our data as shown in Figure 3.

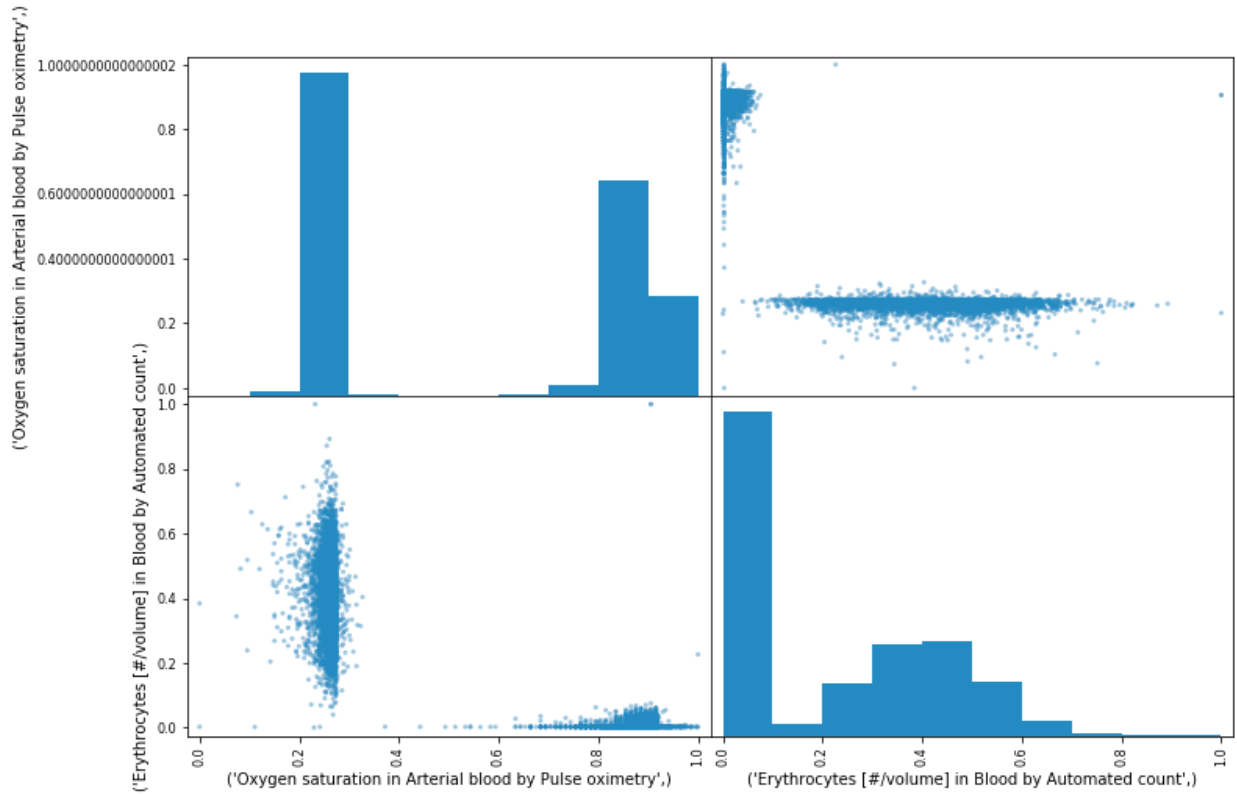

Figure 4: Histograms and Scatter plots of the Two Most Important Features (Oxygen saturation in Arterial blood by Pulse oximetry and Erythrocytes [#/volume] in Blood in Automated count)

From Figure 4 above, we can see the scatter plot between the top two features. From the outcome of the feature selection approach, we can see that the features are represented in two different clusters. Thus, the selected features are also able to separate the clusters properly. Therefore, we did not use Principal Components from our PCA analysis, as the selected features to separate the different clusters.

| Clusters | Avg. Silhouette | Execution Time (s) |
|---|---|---|
| 2 | 0.33251 | 1.51056 |
| 3 | 0.35363 | 1.54716 |
| 4 | 0.36115 | 1.56706 |
| 5 | 0.36254 | 1.55920 |
| 6 | 0.37438 | 1.56353 |
| 7 | 0.38839 | 1.62882 |
| 8 | 0.38791 | 1.62137 |
| 9 | 0.41258 | 1.64178 |
| 10 | 0.41777 | 1.70747 |
| 11 | 0.42404 | 1.70376 |
| 12 | 0.43268 | 1.70742 |
| 13 | 0.43799 | 1.71513 |
| 14 | 0.44313 | 1.73804 |
| 15 | 0.44992 | 1.95996 |
| 16 | 0.45839 | 1.79840 |
| 17 | 0.46399 | 1.81180 |
| 18 | 0.47494 | 1.85525 |
| 19 | 0.47812 | 1.87704 |
| 20 | 0.38621 | 2.09364 |
| 21 | 0.39952 | 1.92338 |
| 22 | 0.40598 | 1.95157 |
| 23 | 0.40791 | 1.97894 |
| 24 | 0.41161 | 2.00387 |
| 25 | 0.41531 | 2.04987 |
| 26 | 0.42146 | 1.97655 |
| 27 | 0.42575 | 2.19522 |
| 28 | 0.40830 | 2.13537 |
| 29 | 0.35308 | 2.16763 |
| 30 | 0.40887 | 2.30241 |

Table 5: Execution Time (s) for Each Cluster Size by Average Silhouette Score in DBSCAN

In Table 5, we have the execution times for each cluster size using the DBSCAN algorithm. We can see that as we increased the cluster size DBSCAN has relatively longer execution times than for smaller cluster sizes. We also see that the average silhouette score tends to increase for higher cluster sizes but inconsistently.

| No. Clusters | Avg. Silhouette Score | min sample # | No. Clusters |
|---|---|---|---|
| 2 | 0.5183527906311406 | 1 | 124 |
| 3 | 0.39076462016119634 | 2 | 6 |
| 4 | 0.2657172251206726 | 3 | 4 |
| 5 | 0.25227461064703033 | 4 | 3 |
| 6 | 0.20455081409550327 | 5 | 2 |
| 7 | 0.17595573653988675 | 6 | 2 |
| 8 | 0.14440748438130696 | 7 | 2 |

Table 6: DBSCAN Number of Clusters by Average Silhouette Score and Minimum Sample Number

| $\varepsilon$ value | No. Clusters | Avg. Silhouette score | Execution Time (s) |
|---|---|---|---|
| 0.1 | 4 | 0.452240591 | 2.557744503 |
| 0.2 | 2 | 0.443145627 | 2.874323130 |
| 0.3 | 2 | 0.543147084 | 3.082974434 |
| 0.4 | 1 | 0.522332234 | 3.120270729 |
| 0.5 | 1 | 0.517065980 | 3.168396950 |
| 0.6 | 1 | 0.518854967 | 3.215951920 |
| 0.7 | 1 | 0.539448619 | 3.454456329 |
| 0.8 | 1 | 0.540967303 | 3.526142597 |
| 0.9 | 1 | 0.540967303 | 3.581864357 |

Table 7: Execution Time (s) for each $\varepsilon$ value with Average Silhouette Score and the Corresponding Number of Clusters in DBSCAN

In Table 6, we have the average silhouette score for each of the clusters of size 2 to 10 and obtain a score greater than 0.5 (0.518356) when the number of clusters is 2 suggesting that a reasonable structure has been found. Additionally, in Table 6, we first obtain the number of clusters to be 2 when the minimum number of samples is 5. Using the minimum number of samples of 5, we can now find the appropriate $\varepsilon$ value corresponding to 2 clusters. In Table 7, we have the execution times for each cluster size using the DBSCAN algorithm with 5 as the minimum number of samples and have determined that the appropriate $\varepsilon$ value should be 0.3 since we obtain an average silhouette score of 0.54314. Lastly, we also note that as we increase the value of $\varepsilon$ we have an increase in execution time for DBSCAN.

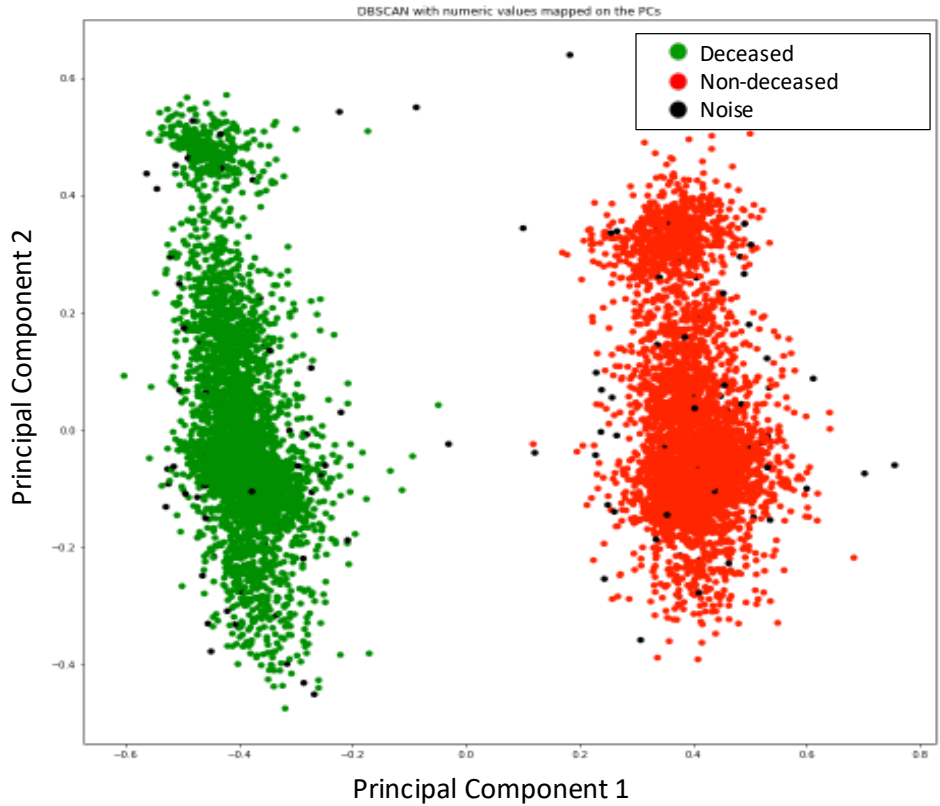

Figure 5: Plot of DBSCAN Clustering of Most Important Numeric Value Features on the first two Principal Components with Red and Green Designating Two Clusters of Core and Border Points and Black Designating Noise Points

For visualizing our data, in Figure 5 we have plotted the labels generated from our DBSCAN given the hyperparameters specified above, minimum number of samples of 5, $\varepsilon = 0.3$, and number of clusters 2, on the principal component 1 vs principal component 2 plane. Note that these points are not the true points in our data rather they are plotted in such a way that we can visualize the points in two dimensions since we have 14-dimensional numeric data points. We see that the two clusters appear to be linearly separatable in this plane and note that DBSCAN has identified noise points in our data that appear to be close to core points and border points at time but only because of how the principal components represent the data points in the plot above.

| $N = 2342$ | $Predicted: Yes$ | $Predicted: No$ | |
|---|---|---|---|
| $Actual: Yes$ | $TP = 1171$ | $FN = 2$ | $SN = 99.83\%$ |
| $Actual: No$ | $FP = 0$ | $TN = 1169$ | $SP = 100\%$ |
| | $Prec = 100\%$ | $Recall = 99.83\%$ | $Accuracy = 99.78\%$ |

Table 8: Confusion Matrix for Support Vector Machine (SVM) for 25% of samples for classifying deceased and non-deceased COVID-19 patients

| $N = 937$ | $Predicted: Yes$ | $Predicted: No$ | |
|---|---|---|---|
| $Actual: Yes$ | $TP = 452$ | $FN = 1$ | $SN = 99.78\%$ |
| $Actual: No$ | $FP = 0$ | $TN = 484$ | $SP = 100\%$ |
| | $Prec = 100\%$ | $Recall = 99.79\%$ | $Accuracy = 100\%$ |

Table 9: Confusion Matrix for Random Forest (RF) [10-Fold Cross Validation] for classifying deceased and non-deceased COVID-19 patients

In Tables 8 and 9, we see the Confusion Matrices for SVM and RF classifier and obtain very high accuracies for both classifiers, accuracy of 99.78% and 100%, respectively. We obtain similar results regarding Sensitivity and Specificity with Sensitivity being 100% for both classifiers and the Specificity being 99.83% and 99.78% for SVM and RF, respectively. Likewise, we obtain 100% precision for both classifiers and 99.83% and 99.97% recall for SVM and RF, respectively. Note that we produced the Confusion Matrix for the SVM classifier with 25% of the data and conducted a 10-Fold Cross Validation for the RF classifier.

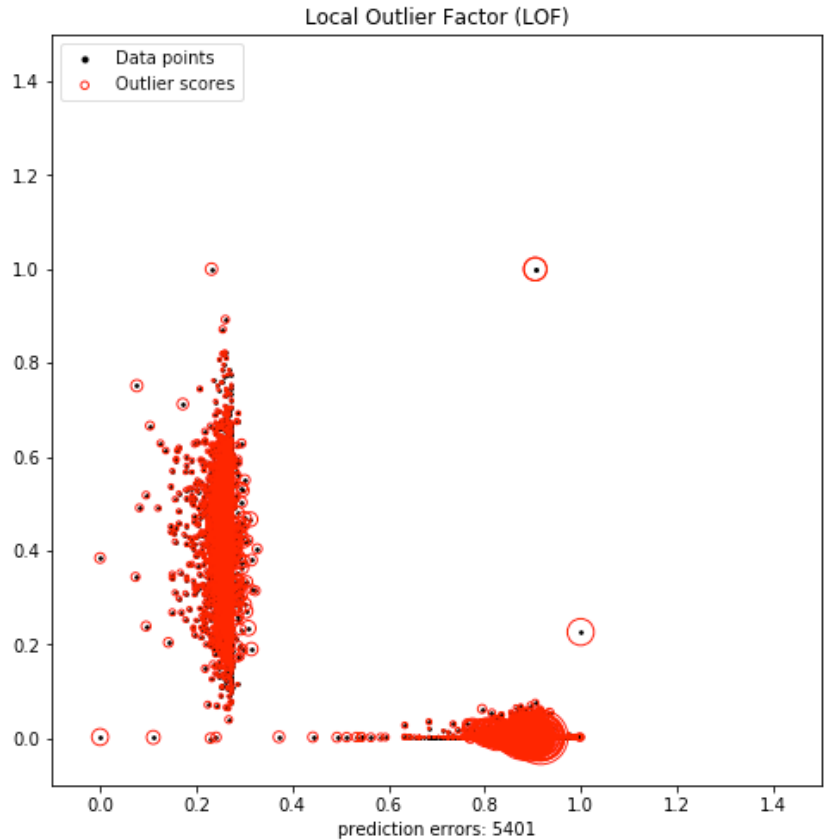

Figure 6: Local Outlier Factor (LOF) Feature Detecting the Outliers in the Dataset

In Figure 6 above, we have the LOF results for detecting outliers of our top two features. We see in the figure that we have multiple outliers as indicated by the points circled red in the plot. The identification of these outlier points depends on the outlier scores generated by the LOF for each point as described above.

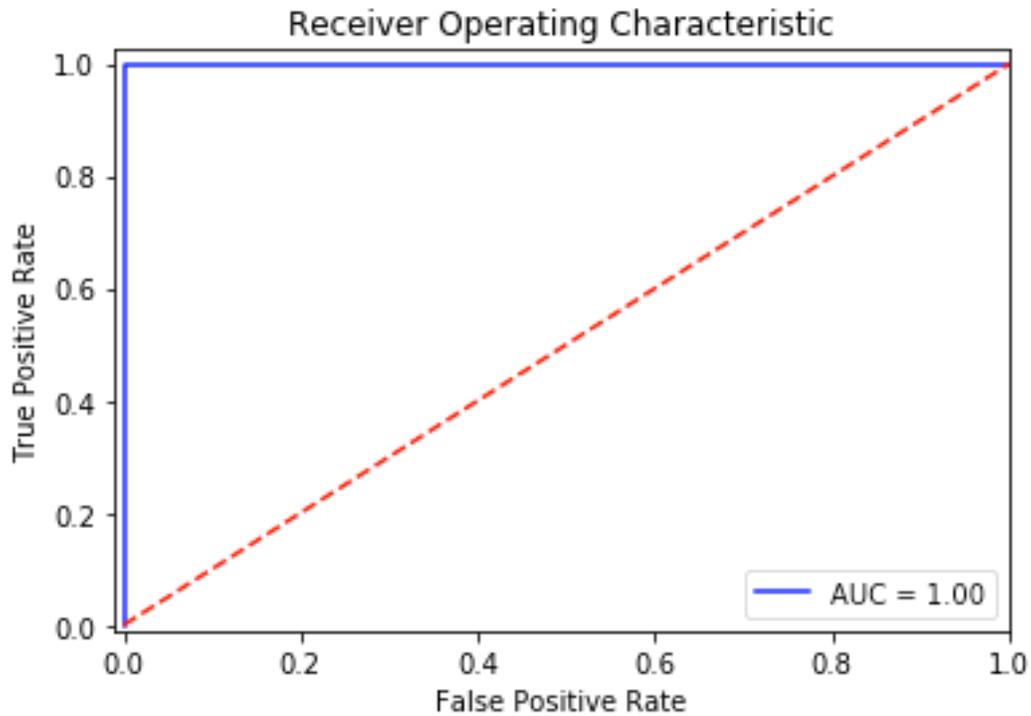

Figure 7: ROC (Receiver Operating Characteristic) curve for Random Forest

In Figure 7, we have the ROC (Receiver Operating Characteristic) curve for Random Forest classifier. Recall from the confusion matrix that we have a near perfect accuracy, ~100%, so it is not surprising that that that we have an $AUC = 1.00$ as shown in this figure. We obtain the same ROC curve for the SVM classifier (*not shown here*).

**Discussion**

The COVID-19 pandemic has affected many lives this year and unfortunately it has caused many deaths for individuals with preexisting condition or comorbidities. In our data mining project, we had the opportunity to interact with healthcare professionals on the frontline caring for patients. We were able to obtain their input and guidance on what patient features might signal or increase the likelihood that COVID-19 patient would not survive. By coordinating with these experts, we were able to identify which attributes to consider for our project and discovered that sepsis and the associated systemic effects of sepsis may signal when a patient is declining and is at high risk of dying from the coronavirus. By including these factors in our study, we found the most important features which classify a patient that has died from a COVID-19 infection.

The top five most important features were Oxygen saturation in Arterial blood by pulse oximetry, Erythrocyte count in Blood, Acute kidney failure, INR, and Severe sepsis with septic shock (Table 3). This is consistent with what physicians would be concerned with in a patient with an acute deadly respiratory viral infection. One would expect that the patient's Oxygen saturation status to correspond with how much the virus has progressed by infecting the lungs decreasing the ability of the lungs to take in oxygen. Similarly, we would expect to have the patient's Erythrocyte count in Blood to increase in attempt to compensate for the lack of oxygen. We also will see end-organ dysfunction in our patients such as Acute kidney failure indicating kidney damage and an increase in INR suggesting that the patient's liver has sustained damage decreasing the amount of coagulation proteins being produced. Also, it is reassuring that we have confirmed that 'Severe sepsis with septic shock' is a valid contributor to our classifier since physicians would have correctly identified these patients as having sepsis as the patient's health status begins to decline ultimately leading to their death. From the other features shown in Table 3, we see addition systemic and end-organ effects consistent with viral septic infections along with relevant patient comorbidities such as Metabolic encephalopathy suggesting metabolic effect due to the lack of oxygen affecting the brain (low oxygen), Acute kidney failure with tubular necrosis (kidney failure), and Acute respiratory distress syndrome (lung involvement).

By conducting a hierarchical clustering, shown in Figure 3, we found that we have two major clusters and at about the level 46 we see that the deceased cluster breaks off into two subclusters followed by the non-deceased cluster breaking off into two subclusters around level 41. It would be interesting to investigate these subclusters further as they may have differing likelihood of dying from COVID-19 in our patient population. In our other clustering method, DBSCAN, we were able to identify the two clusters of interest namely the deceased and non-deceased patients as well as identify the noise points as shown in Figure 5. After tuning the hyperparameters for the DBSCAN clustering we obtained an $\varepsilon = 0.3$ with minimal number of samples of 5 to identify our 2 clusters with the Silhouette coefficient $> 0.5$. Therefore, we have two method which correctly cluster our deceased and non-deceased COVID-19 patients.

From the SVM and RF classification, we were not able to get the results we were expecting and may be concerned by our results that the classifier maybe overfitting the data. We tried several approaches such as feature selection, increasing train size, cross validation to ameliorate this, as well as using other kernels such as the linear, radial basis function, polynomial, sigmoid, and precomputed kernels. However, after visualization we found out that the clusters do not overlap greatly and appear to be easily separable. Therefore, our classifiers and the variants did not have to work very hard to classify the deceased patients from the non-deceased. Thus, providing us with two methods to correctly classify the COVID-19 patients by deceased status. Also, the variable selection approach prevent overfitting but, in our case, after using variable section we still faced an overfitting issue. So, it is possible that variable selection approach may not be suitable for medical data used in our dataset. As there are extreme values among deceased and non-deceased patients which might play a big role for in the variable selection step.

While analyzing the attributes and getting help from our medical professional experts we felt confident that we were picking the right attributes after achieving our results. However, the results do not appear to be novel or new significant findings that our MD colleagues did not already know about the COVID-19 patient population. This poses somewhat of a conformational bias and we may wish to consider conducting a Data Mining project that would discover new relationships that the experts may not have been aware of previously.

Additionally, we should have run our analysis after removing the clear outliers preferably the ones identified in the Appendix, Table 11. However, we did not remove the extreme outliers since these values are consistent with the deceased patient values prior to their death. Also, something that we could have done differently would be to change the resampling method we used. We had a challenging task of sampling as our features that were sparsely populated. We may have preferred to use a Bayesian approach as it would have made sampling for the missing values more reasonable since we do not have normally distributed samples for each of our attributes or features.

Now of we were to do this project again, we would be beginning the IRB process sooner as gaining access to the databases and obtaining approval to work with the data delayed the start of our data cleaning and extraction activities. Unfortunately, we had anticipated spending the Thanksgiving weekend working on our project to find that we could not access the Cerner portal to work on the analysis, also set us back in terms of deadlines and schedules. Also, one of the most challenging limitations to the progress of our project was having to abide by our agreement with Cerner, such that only one of our teammates (Aaron) had access to the data. Meanwhile, Dheeman and Biraj had to use a sanity dataset which had similar attributes and features of the actual dataset and assisted Aaron in the production of the results using the actual dataset. We also faced time constraints since the portal access was limited to 6 AM to 8 PM on Monday through Friday.

We believe that from data in the database we might be able to construct a time series analysis project with the aim of producing a model which would allow for the prediction of non-deceased patients changing status to deceased patients i.e., die from a coronavirus infection. Additionally, determining which patients may need higher levels of care such as going to ICU and being put on a ventilator as these events cost the healthcare system greatly and have the potential to save lives.

Lastly, this project was the first time some of our team members worked with a large raw dataset of this scale. In the data extraction process, we encountered multiple errors and overcame many obstacles and now realize just how much we have learned this semester. In terms of the analysis, we would finally be able to apply the algorithms we learned about in class and discovered their limitations and restrictions when applied to a real dataset. We needed to fix the missing values in our dataset, and we conducted a resampling procedure that none of us had experience with and still feel there is a lot to learn with respect to sampling from a distribution. By experimenting with the sampling procedure through multiple iterations we were able to apply the algorithms to obtain more meaningful results. All in all, this has been a great learning experience and has opened our eyes to the neat and powerful field of Data Mining.

**Conclusion**

Our group members and their respective contributions:

33.33 % - Biraj Tiwari – created plots, tables, and supportive figures, constructed, and ran algorithms, help write documentation, met with clinical and research team to design project

33.34 % - Dheeman Saha – performed data extraction, data analysis, created plots, tables, and supportive figures, constructed and ran algorithms, help write documentation, met with clinical and research team to design project

33.33 % - Aaron Segura - created plots, tables, and supportive figures, help write documentation, submitted IRB and data requests, maintained portal access, assisted in the construction of algorithms and analysis, and met with Drs. Femling and Shekhar to coordinate the construction of the project

# Appendix

Here we have included the additional things we have performed for the data analysis and explanatory purpose.

| All Features Considered | No.NaN |
|---|---|
| 25-Hydroxyvitamin D2+25-Hydroxyvitamin D3 [Mass/volume] in Serum or Plasma | 9045 |
| Alanine aminotransferase [Enzymatic activity/volume] in Serum or Plasma | 4630 |
| Albumin [Mass/volume] in Serum or Plasma | 4015 |
| Alkaline phosphatase [Enzymatic activity/volume] in Serum or Plasma | 1655 |
| Ammonia [Moles/volume] in Plasma | 8548 |
| Aspartate aminotransferase [Enzymatic activity/volume] in Serum or Plasma | 3489 |
| Bilirubin.direct [Mass/volume] in Serum or Plasma | 6139 |
| Bilirubin.total [Mass/volume] in Serum or Plasma | 1833 |
| Body temperature | 4554 |
| C reactive protein [Mass/volume] in Serum or Plasma | 5401 |
| Calcium [Mass/volume] in Serum or Plasma | 1326 |
| Calcium.ionized [Moles/volume] in Blood | 8566 |
| Cholesterol [Mass/volume] in Serum or Plasma | 7307 |
| Cholesterol in HDL [Mass/volume] in Serum or Plasma | 7368 |
| Cholesterol in LDL [Mass/volume] in Serum or Plasma by calculation | 8094 |
| Creatine kinase [Enzymatic activity/volume] in Serum or Plasma | 5988 |
| Creatinine [Mass/volume] in Serum or Plasma | 1265 |
| Diastolic blood pressure | 632 |
| Erythrocytes [#/volume] in Blood by Automated count | 2126 |
| Ferritin [Mass/volume] in Serum or Plasma | 4884 |
| Fibrin D-dimer DDU [Mass/volume] in Platelet poor plasma by Immunoassay | 8735 |
| Fibrin D-dimer FEU [Mass/volume] in Platelet poor plasma | 6663 |
| Heart rate | 7082 |
| Hematocrit [Volume Fraction] of Blood by Automated count | 1358 |
| Hemoglobin A1c/Hemoglobin.total in Blood | 6801 |
| Hemoglobin [Mass/volume] in Blood | 1917 |
| INR in Platelet poor plasma by Coagulation assay | 4407 |
| Influenza virus A Ag [Presence] in Unspecified specimen by Immunoassay | 9366 |
| Influenza virus B Ag [Presence] in Unspecified specimen by Immunoassay | 9366 |
| Inhaled oxygen concentration | 4426 |
| Inhaled oxygen flow rate | 2819 |
| Leukocytes [#/volume] in Blood by Automated count | 3287 |
| Lymphocytes [#/volume] in Blood by Automated count | 2969 |
| MCH [Entitic mass] by Automated count | 1962 |
| MCHC [Mass/volume] by Automated count | 1921 |
| MCV [Entitic volume] by Automated count | 1923 |
| Magnesium [Mass/volume] in Serum or Plasma | 3023 |
| Natriuretic peptide B [Mass/volume] in Blood | 9286 |
| Natriuretic peptide.B prohormone N-Terminal [Mass/volume] in Serum or Plasma | 6796 |
| Neutrophils [#/volume] in Blood by Automated count | 2894 |
| Non-invasive mean arterial pressure | 9050 |
| Oxygen saturation in Arterial blood by Pulse oximetry | 1147 |
| Oxygen therapy | 9359 |
| Oxygen/Inspired gas setting [Volume Fraction] Ventilator | 8809 |
| Platelets [#/volume] in Blood by Automated count | 2262 |
| Potassium [Moles/volume] in Serum or Plasma | 1434 |
| Protein [Mass/volume] in Serum or Plasma | 2026 |
| Prothrombin time (PT) | 3735 |
| Respiratory rate | 90 |
| Respiratory syncytial virus RNA [Presence] in Unspecified specimen by NAA with probe detection | 9366 |
| Sodium [Moles/volume] in Serum or Plasma | 1394 |
| Systolic blood pressure | 271 |
| Triglyceride [Mass/volume] in Serum or Plasma | 6590 |
| Troponin I.cardiac [Mass/volume] in Serum or Plasma | 4498 |
| Troponin T.cardiac [Mass/volume] in Serum or Plasma | 8449 |
| Urate [Mass/volume] in Urine | 9225 |
| Urea nitrogen [Mass/volume] in Serum or Plasma | 1365 |
| aPTT in Platelet poor plasma by Coagulation assay | 5097 |

Table 10: Data for All Features Considered with the Number of Missing Values (No. NaN)

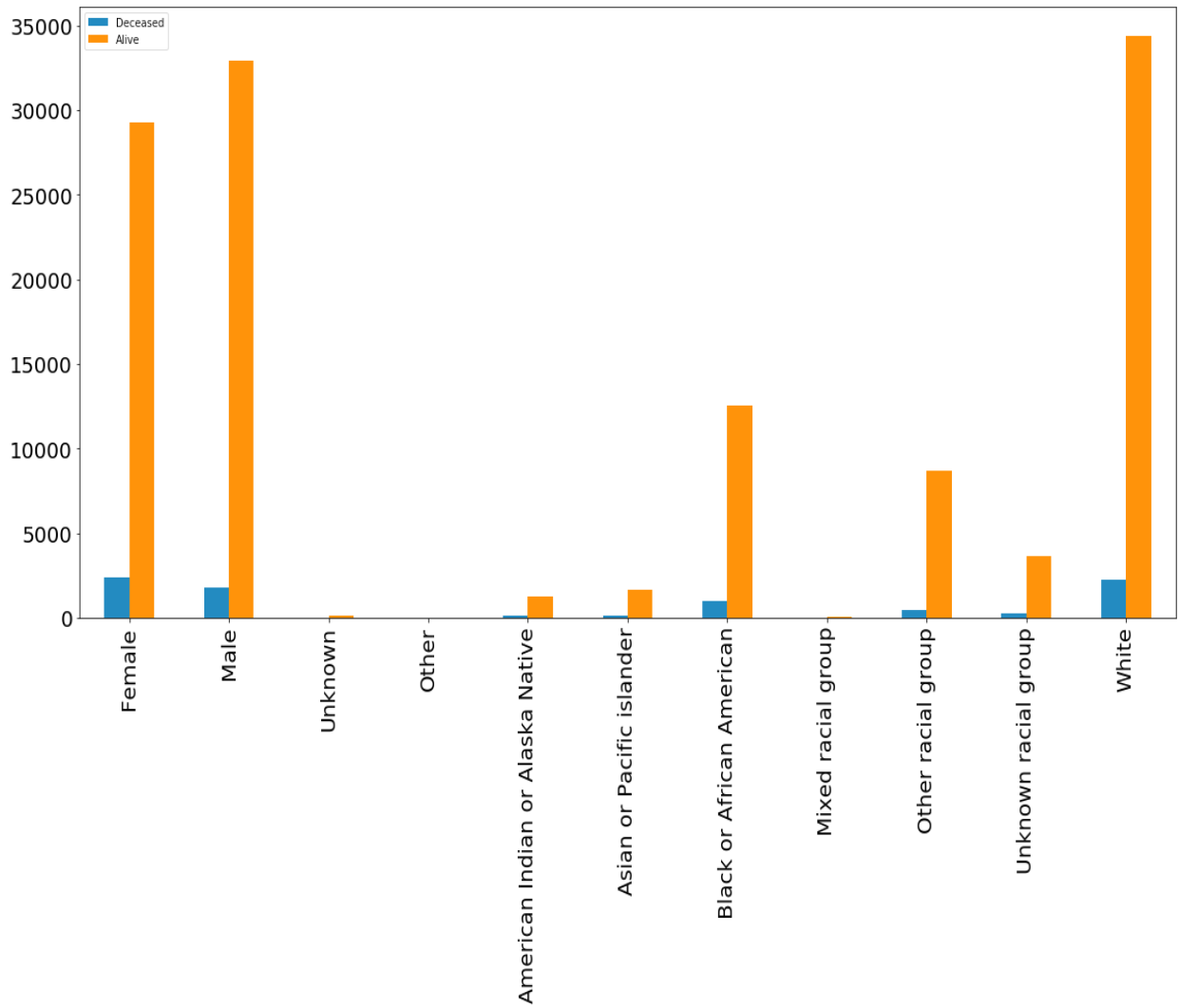

Figure 8: Count of patients separated by gender and Ethnicity grouped by Deceased status in database

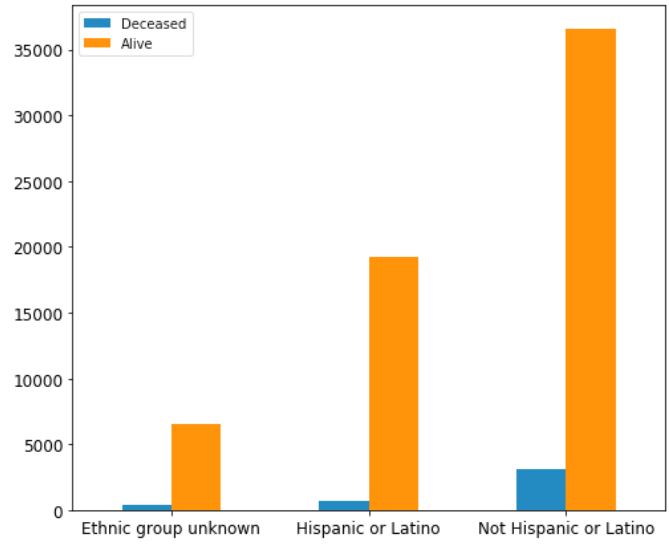

Figure 9: Distribution for Ethnicity grouped by dead/alive

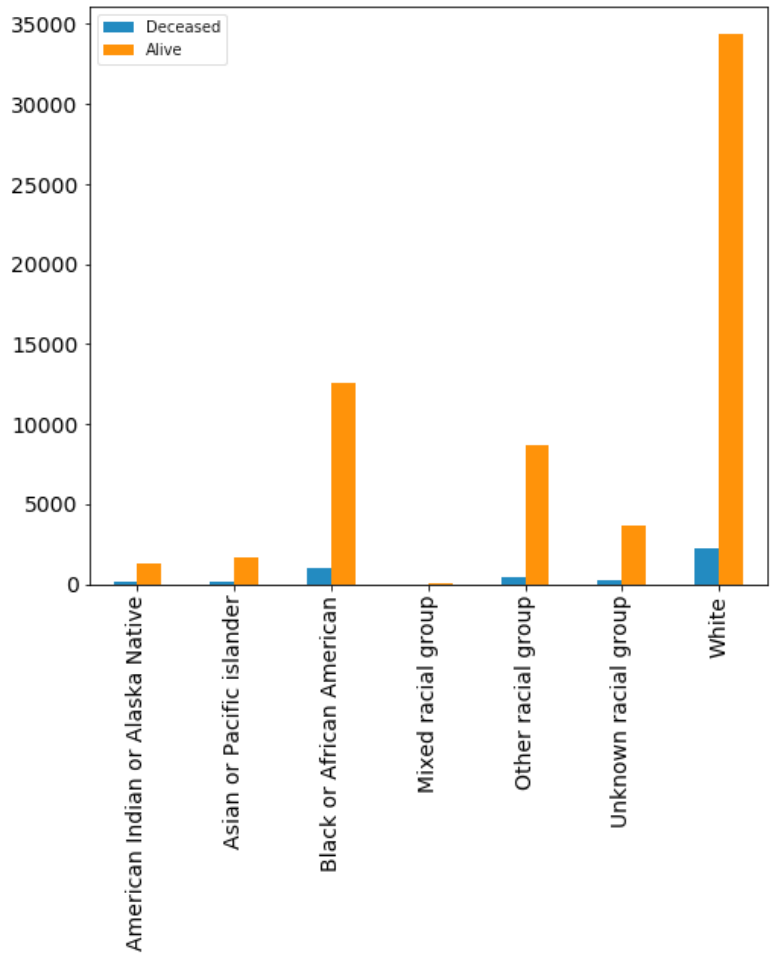

Figure 10: Distribution for race grouped by dead/alive

| Index Column | Number of Outliers | Index Number of Outliers |
|---|---|---|
| 0 | 0 | |
| 1 | 5 | 4689, 4721, 5108, 7623, 7921 |
| 3 | 79 | 69, 146, 221, 315, 413, 478, 755, 790, 986, 1182, 1443,1716, 1749, 2099, 2503, 2523, 2525, 2663, 2798, 2816, 2916, 3027, 3066, 3089, 3178, 3202, 3750, 4194, 4222, 4281, 4297, 4524, 4592, 4662, 4669, 4850, 4929, 4976, 5118, 5145, 5213, 5228, 5297, 5500, 5521, 5782, 6132, 6156, 6184, 6234, 6395, 6481, 6523, 6568, 6601,6633, 6731, 6752, 6777, 6938, 7083, 7358, 7712, 7714, 7822, 7850,7987, 8050, 8086, 8391, 8394, 8410, 8608, 8697, 9026, 9059, 9069,9191, 9263 |
| 5 | 107 | 56, 64, 163, 190, 212, 216, 304, 408, 453, 466, 524 527, 644, 692, 711, 809, 927, 1083, 1235, 1348, 1411, 1465,1519, 1607, 1634, 1645, 1660, 1680, 1700, 1848, 1933, 1961, 2146, 2452, 2787, 2830, 2867, 2907, 3007, 3332, 3370, 3372, 3422, 3424, 3491, 3525, 3641, 3759, 4026, 4087, 4127, 4156, 4240, 4322, 4502, 4607, 4619, 4772, 4832, 4960, 5000, 5006, 5086, 5115, 5339, 5430, 5497, 5582, 5851, 5903, 5925, 6183, 6411, 6860, 6933, 6966, 7129, 7261, 7334, 7345, 7429, 7462, 7471, 7532, 7772, 7821, 7921, 7930,7991, 8040, 8162, 8224, 8512, 8706, 8744, 8765, 8825, 8924, 8934, 8968, 8991, 8995, 9039, 9065, 9108, 9164, 9332 |
| 6 | 98 | 33, 213, 258, 524, 529, 549, 609, 637, 664, 847, 952,1067, 1083, 1138, 1675, 1813, 1939, 2106, 2327, 2331, 2559, 2744, 2888, 2952, 2968, 3071, 3333, 3418, 3558, 3579, 3604, 3641, 3759, 3834, 3908, 4087, 4090, 4091, 4221, 4244, 4308, 4333, 4343, 4378, 4477, 4607, 4694, 4828, 4924, 4986, 5194, 5293, 5441, 5467, 5591, 5768, 5829, 5949, 5977, 5991, 6128, 6267, 6467, 6533, 6613, 6638, 6716, 6744, 6866, 6924, 6943, 7300, 7562, 7590, 7636, 7691, 7740, 7772, 8175, 8208, 8253, 8496, 8513, 8547, 8576, 8577, 8590, 8706,8751, 8825, 8864, 8940, 8971, 9055, 9135, 9251, 9290, 9312 |
| 8 | 23 | 276, 356, 1366, 1561, 1608, 1834, 1857, 1872, 2175, 2536, 2644, 4607, 4610, 6788, 7061, 7238, 7507, 7567, 7921, 7929, 8596, 9253 |
| 9 | 84 | 183, 337, 410, 413, 468, 488, 649, 759, 954,1035,1253,1307, 1344, 1361, 1489, 1523, 1562, 1672, 1766, 1864, 1879, 2004, 2111, 2145, 2233, 2256, 2380, 2463, 2887, 2991, 3078, 3155, 3169, 3184, 3196, 3233, 3246, 3264, 3295, 3375, 3416, 3638, 3672, 3791, 3824, 3851, 3907, 4402, 4488, 4755, 5077, 5087, 5523, 5777, 5859,6057, 6081, 6100, 6162, 6183, 6204, 6356, 6481, 6493, 6547, 6649, 6652, 6721, 6916, 6920, 6931, 7275, 7418, 7834, 7881, 7892, 8023  8342, 8522, 8624, 8653, 8858, 8864, 8963 |
| 10 | 43 | 62, 211, 357, 427, 435, 517, 577, 689, 767, 829, 941, 1174, 1461, 1549, 1620, 1833, 2344, 2409, 2536, 2962, 3014, 3696, 4444, 4450, 4562, 4716, 4724, 5006, 5160, 5215, 5692, 5781, 5894, 6097, 6341, 6555, 7023, 7216, 7701, 8159, 8991, 9159, 9360] |
| 11 | 79 | 145, 169, 200, 601, 655, 681, 837, 915, 954,1166,1300,1360, 1378, 1522, 1834, 1924, 1950, 1978, 2035, 2160, 2596, 2794, 2801, 2830, 3192, 3483, 3495, 3525, 3566, 4123, 4363, 4490, 4494, 4598, 4912, 5037, 5043, 5099, 5106, 5243, 5300, 5534, 5562, 5567, 5742, 5768, 5789, 5819, 6026, 6064, 6128, 6182, 6184, 6633, 6766, 6788, 6937, 7032, 7072, 7248, 7275, 7580, 7696, 7806, 7822, 7897, 7939, 8127, 8135, 8317, 8373, 8447, 8526, 8884, 9082, 9227, 9249, 9285, 9290 |
| 13 | 19 | 62, 276, 356, 947, 1608, 1834, 1857, 2213, 2536, 4272, 5829, 6788, 6826, 7507, 7567, 7809, 7929, 8596, 9035 |
| 14 | 57 | 281, 324, 374, 596, 607, 631, 962, 1029, 1250, 1449, 1472,1521, 1736, 1804, 1848, 1913, 2191, 2373, 2379, 2527, 2691, 2803, 3083, 3276, 3366, 3595, 4404, 4504, 4524, 4554, 4568, 4859, 4976, 5294, 5583, 5681, 6072, 6113, 6204, 6565, 6567, 6671, 6756, 6771, 7317, 7361, 7660, 7826, 7859, 8121, 8207, 8336, 8531, 9188, 9296, 9308, 9344 |
| 15 | 115 | 48, 123, 286, 343, 411, 657, 690, 695, 711, 816, 965, 1011, 1086, 1191, 1247, 1334, 1378, 1415, 1575, 1816, 1824, 1943,1992, 2043, 2058, 2106, 2141, 2389, 2531, 2534, 2595, 2637, 2789, 2864, 2911, 2925, 2945, 3054, 3188, 3198, 3205, 3246, 3300, 3304, 3546, 3610, 3686, 3902, 3994, 4205, 4287, 4322, 4371, 4387, 4554, 4663, 4716, 4868, 4984, 5101, 5221, 5362, 5474, 5576, 5591, 5648, 5782, 5802, 5933, 6042, 6057, 6083, 6113, 6226, 6245, 6299, 6301, 6413, 6632, 6656, 6686, 6829, 6874, 6967, 6986, 7059, 7140, 7308,7411, 7430, 7454, 7462, 7771, 7798, 7860, 7991, 8043, 8199, 8224, 8282, 8390, 8415, 8518, 8609, 8610, 8620, 8713, 8739, 8822, 8824, 8933, 8938, 8991, 9236, 9357 |
| 17 | 98 | 8, 102, 144, 409, 410, 413, 649, 668, 791, 846, 954, 961, 963, 1006, 1035,1085, 1300, 1318, 1553, 1987, 2112, 2145, 2316, 2358, 2380, 2456, 2596, 2658, 2691, 2830, 2937, 2955, 3005,3107, 3169, 3196, 3246, 3373, 3416, 3464, 3679, 3682, 4245, 4411, 4457, 4557, 4662, 4755, 4769, 4882, 5001, 5075, 5475, 5508, 5729, 5827, 5848, 5859, 5930, 6002, 6057, 6204, 6391, 6481, 6547, 6649, 6652, 6704, 6875, 6916, 6931, 6944, 7175, 7188, 7275, 7363, 7418,7482, 7634, 7638, 7671, 7779, 7834, 7881, 7949, 8023, 8126, 8342,8522, 8624, 8649, 8653, 8788, 8858, 9054, 9098, 9148, 9249 |
| 18 | 70 | 29, 96, 216, 313, 593, 621, 817, 938, 960, 1381, 1405, 1470, 1508, 1682, 1775, 1834, 1903, 1907, 2127, 2134, 2318, 2344, 2540, 2778, 3271, 3304, 3383, 3394, 3725, 3730, 3791, 4221, 4391,4456, 4537, 4615, 4718, 4784, 4870, 5006, 5063, 5076, 5099, 5125,5162, 5439, 5586, 5613, 5669, 5894, 6086, 6107, 6182, 6328, 6590,6613, 6669, 7000, 7028, 7471, 7537, 7554, 7847, 8165, 8243, 8539,8674, 8778, 8929, 9212 |

Table 11: Local Outlier Factor (LOF) by Index column with the Number of Outliers and their Index Number

| Patient ID | Oxygen saturation in Arterial blood by Pulse oximetry | Erythrocytes [#/volume] in Blood by Automated count | Acute kidney failure, unspecified | INR in Platelet poor plasma by Coagulation assay | Severe sepsis with septic shock | Sodium [Moles/volume] in Serum or Plasma | Magnesium [Mass/volume] in Serum or Plasma | Cardiac arrest, cause unspecified | Hematocrit [Volume Fraction] of Blood by Automated count,) | Aspartate aminotransferase [Enzymatic activity/volume] in Serum or Plasma | MCHC [Mass/volume] by Automated count | Potassium [Moles/volume] in Serum or Plasma | Acute kidney failure with tubular necrosis | Hemoglobin [Mass/volume] in Blood | Body temperature | Respiratory rate | Acute respiratory distress syndrome | Alanine aminotransferase [Enzymatic activity/volume] in Serum or Plasma | Platelets [#/volume] in Blood by Automated count | Metabolic encephalopathy |
|---|---|---|---|---|---|---|---|---|---|---|---|---|---|---|---|---|---|---|---|---|
|  | 0.26502 | 0.31983 | 1 | 0.01677 | 0 | 0.49450 | 0.30000 | 0 | 0.16197 | 0.00114 | 0.89067 | 0.15779 | 0 | 0.25665 | 0.25345 | 0.21359 | 0 | 0.00219 | 0.21820 | 1 |
| 1 | 0.78478 | 0.00209 | 0 | 0.07958 | 0 | 0.40625 | 0.17787 | 0 | 0.45265 | 0.01802 | 0.69721 | 0.30000 | 0 | 0.44632 | 0.14800 | 0.26315 | 0 | 0.00861 | 0.22085 | 0 |
| 2 | 0.26502 | 0.56564 | 1 | 0.02229 | 1 | 0.52747 | 0.34000 | 0 | 0.52640 | 0.01870 | 0.81028 | 0.26306 | 0 | 0.52141 | 0.25139 | 0.19417 | 1 | 0.00021 | 0.15921 | 1 |
| 3 | 0.24316 | 0.45487 | 1 | 0.04509 | 1 | 0.67033 | 0.46000 | 0 | 0.41197 | 0.00042 | 0.80444 | 0.20458 | 0 | 0.24908 | 0.18332 | 0.27184 | 1 | 0.00081 | 0.12287 | 1 |
| 4 | 0.29954 | 0.37709 | 1 | 0.01502 | 0 | 0.57142 | 0.26000 | 0 | 0.48943 | 0.00360 | 0.79742 | 0.21627 | 0 | 0.40430 | 0.28776 | 0.17475 | 0 | 0.00067 | 0.10324 | 0 |
| ... | ... | ... | ... | ... | ... | ... | ... | ... | ... | ... | ... | ... | ... | ... | ... | ... | ... | ... | ... | ... |
| 9361 | 0.26776 | 0.49206 | 0 | 0.07110 | 0 | 0.57142 | 0.51427 | 0 | 0.34652 | 0.00027 | 0.86346 | 0.23967 | 0 | 0.55705 | 0.63320 | 0.21359 | 0 | 0.00067 | 0.00968 | 0 |
| 9362 | 0.23770 | 0.60820 | 0 | 0.01044 | 0 | 0.41758 | 0.38000 | 0 | 0.41021 | 0.00048 | 0.76172 | 0.16948 | 1 | 0.44985 | 0.00811 | 0.19417 | 0 | 0.00175 | 0.21699 | 0 |
| 9363 | 0.91558 | 0.00154 | 1 | 0.11547 | 0 | 0.46875 | 0.32494 | 0 | 0.34003 | 0.01373 | 0.760956 | 0.38333 | 0 | 0.36158 | 0.41660 | 0.13157 | 0 | 0.01245 | 0.11309 | 0 |
| 9364 | 0.86527 | 0.00133 | 0 | 0.07998 | 0 | 0.42912 | 0.27262 | 0 | 0.28075 | 0.08940 | 0.67330 | 0.30837 | 0 | 0.27118 | 0.34729 | 0.13157 | 0 | 0.09110 | 0.16933 | 1 |
| 9365 | 0.26458 | 0.30865 | 1 | 0.01627 | 1 | 0.64835 | 0.34000 | 0 | 0.16901 | 0.00900 | 0.73311 | 0.27476 | 0 | 0.23120 | 0.26444 | 0.19417 | 0 | 0.01775 | 0.24287 | 0 |

Table 12: Sample of Top 20 Most Important Features for Predicting Death in COVID-19 Patient Data

|  | Oxygen saturation in Arterial blood by Pulse oximetry | Erythrocytes [#/volume] in Blood by Automated count | Acute kidney failure, unspecified | INR in Platelet poor plasma by Coagulation assay | Severe sepsis with septic shock | Sodium [Moles/volume] in Serum or Plasma | Magnesium [Mass/volume] in Serum or Plasma | Cardiac arrest, cause unspecified | Hematocrit [Volume Fraction] of Blood by Automated count,) | Aspartate aminotransferase [Enzymatic activity/volume] in Serum or Plasma | MCHC [Mass/volume] by Automated count | Potassium [Moles/volume] in Serum or Plasma | Acute kidney failure with tubular necrosis | Hemoglobin [Mass/volume] in Blood | Body temperature | Respiratory rate | Acute respiratory distress syndrome | Alanine aminotransferase [Enzymatic activity/volume] in Serum or Plasma | Platelets [#/volume] in Blood by Automated count | Metabolic encephalopathy |
|---|---|---|---|---|---|---|---|---|---|---|---|---|---|---|---|---|---|---|---|---|
| count | 9366 | 9366 | 9366 | 9366 | 9366 | 9366 | 9366 | 9366 | 9366 | 9366 | 9366 | 9366 | 9366 | 9366 | 9366 | 9366 | 9366 | 9366 | 9366 | 9366 |
| mean | 0.56868 | 0.20567 | 0.43583 | 0.05663 | 0.20414 | 0.49308 | 0.30139 | 0.09641 | 0.46487 | 0.03009 | 0.72945 | 0.31511 | 0.11477 | 0.45942 | 0.38212 | 0.18196 | 0.09822 | 0.03284 | 0.15837 | 0.17456 |
| std | 0.31234 | 0.21724 | 0.49582 | 0.05719 | 0.40309 | 0.09190 | 0.09533 | 0.29517 | 0.14264 | 0.04646 | 0.07369 | 0.10391 | 0.31876 | 0.13636 | 0.18097 | 0.07609 | 0.29763 | 0.04421 | 0.07651 | 0.37961 |
| min | 0 | 0 | 0 | 0 | 0 | 0 | 0 | 0 | 0 | 0 | 0 | 0 | 0 | 0 | 0 | 0 | 0 | 0 | 0 | 0 |
| 25% | 0.26229 | 0.00233 | 0.00000 | 0.01500 | 0 | 0.43750 | 0.23525 | 0 | 0.36583 | 0.00171 | 0.68529 | 0.23962 | 0.00000 | 0.36357 | 0.25277 | 0.13157 | 0.00000 | 0.00385 | 0.10565 | 0 |
| 50% | 0.31975 | 0.06813 | 0.00000 | 0.03872 | 0 | 0.48437 | 0.28000 | 0 | 0.47183 | 0.01716 | 0.73311 | 0.30986 | 0.00000 | 0.46327 | 0.34079 | 0.16504 | 0.00000 | 0.01918 | 0.15030 | 0 |
| 75% | 0.88539 | 0.40537 | 1.00000 | 0.07814 | 0 | 0.54185 | 0.36000 | 0 | 0.56726 | 0.04089 | 0.78130 | 0.37217 | 0.00000 | 0.55705 | 0.46155 | 0.21052 | 0.00000 | 0.04651 | 0.20144 | 0 |
| max | 1 | 1 | 1 | 1 | 1 | 1 | 1 | 1 | 1 | 1 | 1 | 1 | 1 | 1 | 1 | 1 | 1 | 1 | 1 | 1 |

Table 13: Summary Statistics of Top 20 Most Important Features for Predicting Death in COVID-19 Patient Data